\RequirePackage{fix-cm}
\documentclass[smallcondensed]{svjour3}     
\smartqed  
\usepackage{graphicx}

\usepackage{amsmath}
\usepackage{amsfonts}

\usepackage[ruled,vlined]{algorithm2e}

\usepackage{subfigure}
\usepackage{subfigmat}

\begin{document}

\title{Reinforcement Learning for Robust Missile Autopilot Design
}


\author{Bernardo Cortez
\and
        Florian Peter
\and
        Thomas Lausenhammer
\and
        Paulo Oliveira
}


\institute{\emph{F. Author} \\
Técnico Lisboa \and \email{bernardo.g.f.cortez@tecnico.ulisboa.pt \and
ORCID: 0000-0001-5553-1106}           
           \\
           \emph{S. Authors}    \\
              MBDA Deutschland GmbH \and \email{florian.peter@mbda-systems.de}
            \and    \\
              Technical University of Munich \and \email{thomas.lausenhammer@tum.de}
            \and    \\
              Tecnico Lisboa \and \email{paulo.j.oliveira@tecnico.ulisboa.pt}
}

\date{Accepted: August 10th 2021}

\maketitle

\begin{abstract}

Designing missiles’ autopilot controllers has been a complex task, given the extensive flight envelope and the nonlinear flight dynamics. A solution that can excel both in nominal performance and in robustness to uncertainties is still to be found.  While Control Theory often debouches into parameters’ scheduling procedures, Reinforcement Learning has presented interesting results in ever more complex tasks, going from videogames to robotic tasks with continuous action domains. However, it still lacks clearer insights on how to find adequate reward functions and exploration strategies. To the best of our knowledge, this work is pioneer in proposing Reinforcement Learning as a framework for flight control. In fact, it aims at training a model-free agent that can control the longitudinal non-linear flight dynamics of a missile, achieving the target performance and robustness to uncertainties. To that end, under TRPO’s methodology, the collected experience is augmented according to HER, stored in a replay buffer and sampled according to its significance. Not only does this work enhance the concept of prioritized experience replay into BPER, but it also reformulates HER, activating them both only when the training progress converges to suboptimal policies, in what is proposed as the SER methodology. The  results  show  that  it  is  possible  both  to  achieve  the  target performance  and to  improve  the agent’s robustness to uncertainties (with low damage on nominal performance) by further training it in non-nominal environments, therefore validating the proposed approach and encouraging future research in this field.

\end{abstract}









\section{Introduction}
\label{sec:intro}

Over the last decades, designing the autopilot flight controller for a system such as a missile has been a complex task, given (i) the non-linear dynamics and (ii) the demanding performance and robustness requirements. This process is classically solved by tedious scheduling procedures, which often lack the ability to generalize across the whole flight envelope and different missile configurations.

Reinforcement Learning (RL) constitutes a promising approach to address the latter issues, given its ability of controlling systems from which it has no prior information. This motivation increases as the system to be controlled grows in complexity, for the plausible hypothesis that the RL agent could benefit from having a considerably wider range of possible (combinations of) actions. By learning the cross-coupling effects, the agent is expected to converge to the optimal policy faster.

The longitudinal nonlinear dynamics (cf. section \ref{sec:model}), however, constitutes a mere first step, necessary to the posterior possible expansion of the approach to the whole flight dynamics. This work is, hence, motivated by the will of finding a RL algorithm that can control the longitudinal nonliear flight dynamics of a Generic Surface-to-Air Missile (GSAM) with no prior information about it, being, thus, model-free.

\section{Model}
\label{sec:model}

\subsection{GSAM Dynamics}
\label{section:missile_dynamics}

The GSAM's flight dynamics to be controlled is modelled by a nonlinear system, in which the total sum of the forces $F_{T,G}$ and of the momenta $M_{T,G}$ are dependent on the flight dynamic states (cf. equations \eqref{eq:dynamic_force} and \eqref{eq:dynamic_momenta}), like the Mach number $M$, the height $h$ or the angle of attack $\alpha$.

\begin{gather}
    F_{T,G} = F(M, h, \alpha, ...) \label{eq:dynamic_force}    \\
    M_{T,G} = M(M, h, \alpha, ...) \label{eq:dynamic_momenta}
\end{gather}

This system is decomposed into Translation (cf. equation \eqref{eq:v_diff}), Rotation (cf. equation \eqref{eq:omega_diff}), Position (cf. equation \eqref{eq:pos_diff}) and Attitude (cf. equation \eqref{eq:att_diff}) terms, as Peter (2018) \cite{Peter2018} described. Equations \eqref{eq:v_diff} to \eqref{eq:att_diff} follow Peter's (2018) \cite{Peter2018} notation.

\begin{gather}
    \dot{V}_G^E=\frac{1}{m}F_{T,G}-\omega^E\times V_G^E    \label{eq:v_diff}  \\
    \dot{\omega}^E = \left(J_G\right)^{-1}\left( M_{T,G}-\omega^E\times J_G\omega^E\right)\label{eq:omega_diff} \\
    \dot{r}_G^E = M_TV_G^E   \label{eq:pos_diff}    \\
    \begin{bmatrix}
        \dot{\Phi}^E, \dot{\Theta}^E, \dot{\Psi}^E
    \end{bmatrix}^T =
    R.\omega^E   \label{eq:att_diff}
\end{gather}

To investigate the principle ability of an RL agent to serve as a missile autopilot, the nonlinear longitudinal motion of the missile dynamics is considered. Therefore, the issue of cross-coupling effects within the autopilot design is not addressed in this paper.

\subsection{Actuator Dynamics}
\label{section:actuator}

The GSAM is actuated by four fin deflections, $\delta_i$, which are mapped to three aerodynamic equivalent controls, $\xi$, $\eta$ and $\zeta$ (cf. equation \eqref{eq:fin2aero}). The latter provide the advantage of directly matching the GSAM's roll, pitch and yaw axis, respectively.

\begin{gather}
    \begin{bmatrix}
        \xi    \\
        \eta    \\
        \zeta
    \end{bmatrix} = \frac{1}{4} \begin{bmatrix}
                                    1 & 1 & 1 & 1 \\
                                    1 & -1 & -1 & 1 \\
                                    1 & 1 & -1 & -1
                                \end{bmatrix}   \begin{bmatrix}
                                                    \delta_1    \\
                                                    \delta_2    \\
                                                    \delta_3    \\
                                                    \delta_4
                                                \end{bmatrix}   \label{eq:fin2aero}   \\
    \begin{bmatrix}
        \delta_1    \\
        \delta_2    \\
        \delta_3    \\
        \delta_4
    \end{bmatrix} = \begin{bmatrix}
                                    1 & 1 & 1 \\
                                    1 & -1 & 1 \\
                                    1 & -1 & -1 \\
                                    1 & 1 & -1
                                \end{bmatrix}   \begin{bmatrix}
                                                    \xi    \\
                                                    \eta    \\
                                                    \zeta
                                                \end{bmatrix}   \label{eq:aero2fin}
\end{gather}

In the control of the longitudinal flight dynamics, $\xi$ and $\zeta$ are set to 0 and, hence, equation \eqref{eq:aero2fin} becomes equation \eqref{eq:longitudinal_actuator}.

\begin{equation}
    \begin{bmatrix}
        \delta_1    \\
        \delta_2    \\
        \delta_3    \\
        \delta_4
    \end{bmatrix} = \begin{bmatrix}
                                    1 \\
                                    -1 \\
                                    -1 \\
                                    1
                                \end{bmatrix}\eta   \label{eq:longitudinal_actuator}
\end{equation}

The actuator system is, thus, the system that receives the desired (commanded) $\eta_{com}$ and outputs $\eta$, modelling the dynamic response of the physical fins with its deflection limit of 30 degrees. The latter is assumed to be a second order system with the following closed loop characteristics:
\begin{enumerate}
    \item Natural frequency $\omega_n$ of 150 rad.s$^{-1}$
    \item Damping factor $\lambda$ of 0.7
\end{enumerate}

\subsection{Performance Requirements}
\label{section:requirements}

The algorithm must achieve the target performance established in terms of the following requirements:

\begin{enumerate}
    \item Static error margin = 0.5\%   \label{item:static_error_margin}
    \item Overshoot $<$ 20\%    \label{item:overshoot}
    \item Rise time $<$ 0.6s \label{item:rise_time}
    \item Settling time (5\%) $<$ 0.6s    \label{item:sett_time}
    \item Bounded actuation
    \item Smooth actuation
\end{enumerate}

Besides, the present work also aims at achieving and improving the robustness of the algorithm to conditions different from the training ones.

When trying to optimize the achieved performance, one must, however, be aware of the physical limitations imposed by the system being controlled. A commercial airplane, for example, will never have the agility of a fighter jet, regardless of how optimized its controller is. Therefore, it is hopeless to expect the system to follow too abrupt reference signals like step functions. Instead, the system will be asked to follow \textit{shaped} reference signals, i.e., the output of the Reference Model. The latter is hereby defined as the system whose closed loop dynamics is designed to mimic the one desired for the dynamic system being controlled. In this case: natural frequency $\omega_n$ and damping factor $\lambda$ of 10 rad.s$^{-1}$ and 0.7, respectively.

\begin{figure}[!htb]
  \centering
  \includegraphics[width=1\textwidth]{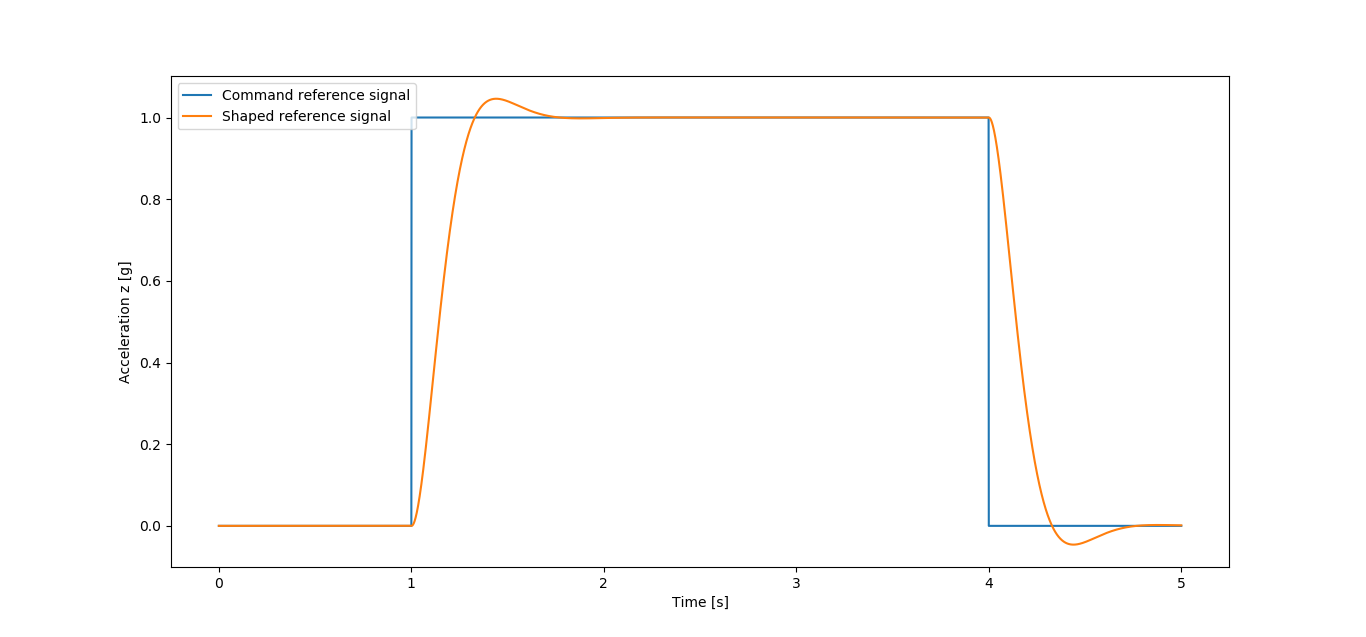}
  \caption{Shaped and command reference signals}
  \label{fig:command_shaped_ref}
\end{figure}

\begin{figure}[!htb]
  \centering
  \includegraphics[width=1\textwidth]{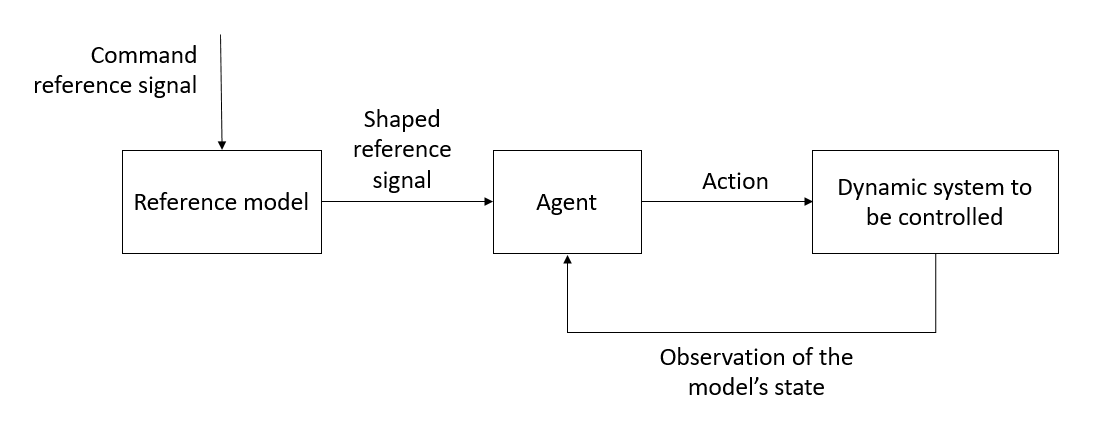}
  \caption{Block diagram including the Reference Model}
  \label{fig:ref_model_diagram}
\end{figure}

The proper workflow of the interaction between the agent and the dynamic system being controlled - including a Reference Model - is illustrated in figure \ref{fig:ref_model_diagram}. The \textit{command} reference signal is given as an input to the Reference Model, whose output, the \textit{shaped} reference signal is the input of the agent.

\section{Background}
\label{sec:backg}

\subsection{Topic Overview}
\label{sec:topic_overview}

RL has been the object of research after the foundations laid out by Sutton et al. (2018) \cite{Sutton2018}, with applications going from the Atari 2600 games, to the MuJoCo games, or to robotic tasks (like grasping...) or other classic control problems (like the inverted pendulum).

TRPO (Schulman et al. 2015 \cite{Schulman2015}) has become one of the commonly accepted benchmarks for its success achieved by the cautious trust region optimization and monotonic reward increase. Perpendicularly, Lillicrap et al. (2016) \cite{Lillicrap2016} proposed DDPG, a model-free off-policy algorithm, revolutionary not only for its ability of learning directly from pixels while maintaining the network architecture simple, but mainly because it was designed to cope with continuous domain action. Contrarily to TRPO, DDPG's off-policy nature implied a much higher sample efficiency, resulting in a faster training proccess. Both TRPO and DDPG have been the roots for much of the research work that followed.

On the one side, some authors valued more the benefits of an off-policy algorithm and took the inspiration in DDPG to develop TD3 (Fujimoto et al. 2018 \cite{Fujimoto2018}), addressing DDPG's problem of over-estimation of the states' value. On the other side, others preferred the benefits of on-policy algorithm and proposed interesting improvements to the original TRPO, either by reducing its implementation complexity (Wu et al. 2017 \cite{Wu2017}), by trying to decrease the variance of its estimates (Schulman 2016 \cite{Schulman2016a}) or even by showing the benefits of its interconnection with replay buffers (Kangin et al. \cite{Kangin2019}).

Apart from these, a new result began to arise: agents were ensuring stability at the expense of converging to suboptimal solutions. Once again, new algorithms were conceived in each family, on- and off-policy. Haarnoja and Tang proposed to express the optimal policy via a Boltzmann distribution in order to learn stochastic behaviors and to improve the exploration phase within the scope of an off-policy actor-critic architecture: Soft Q-learning (Haarnoja et al. 2017 \cite{Haarnoja2017}). Almost simultaneously, Schulman et al. published PPO (Schulman et al. 2017 \cite{Schulman2017}), claiming to have simplified TRPO's implementation and increased its sample-efficiency by inserting an entropy bonus to increase exploration performance and avoid premature suboptimal convergence. Furthermore, Haarnoja et al. developed SAC (Haarnoja et al. 2018 \cite{Haarnoja2018}), in an attempt to recover the training stability without losing the entropy-encouraged exploration and Nachum et al. (2018) \cite{Nachum2018a} proposed Smoothie, allying the trust region implementation of PPO with DDPG.

Finally, there has also been research done on merging both on-policy and off-policy algorithms, trying to profit from the upsides of both, like IPG (Gu et al. 2017 \cite{Gu2017}), TPCL (Nachum et al. 2018 \cite{Nachum2018}), Q-Prop (Guet al. 2019 \cite{Gu2019}) and PGQL (O'Donoghue et al. 2019 \cite{ODonoghue2019}).

\subsection{Trust Region Policy Optimization}\label{sec:trpo}

TRPO is an on-policy model-free RL algorithm that aims at maximizing the discounted sum of future rewards (cf. equation \eqref{eq:sum_disc_fut_rew}) following an actor-critic architecture and a trust region search.

\begin{equation}
    R(s_t) = \sum^\infty_{l=0}\gamma^lr(t+l)
    \label{eq:sum_disc_fut_rew}
\end{equation}

Initially, Schulman et al. (2015) \cite{Schulman2015} proposed to update the policy estimator's parameters with the conjugate gradient algorithm followed by a line search. The trust region search would be ensured by a hard constraint on the Kullback-Leibler divergence $D_{KL}$.

Briefly after, Kangin et al. (2019) \cite{Kangin2019} proposed an enhancement, augmenting the training data by using replay buffers and  GAE (Schulman et al. 2016 \cite{Schulman2016a}). Also contrarily to the original proposal, Kangin et al. trained the value estimator's parameters with the ADAM optimizer and the policy's with K-FAC (Martens et al. 2015 \cite{Martens2015}). The former was implemented within a regression between the output of the Value NN, $\hat{V}$, and its target, $V'$, whilst the latter used equation \eqref{eq:pol_loss_paper} as the loss function, which has got a first term concerning the objective function being maximized (cf. equation \eqref{eq:sum_disc_fut_rew}) and a second one penalizing differences between two consecutive policies outside the trust region, whose radius is the hyperparameter $\delta_{TR}$.

\begin{gather}
    \begin{split}\label{eq:pol_loss_paper}
        L_P = -\mathbb{E}_{s_0,a_0,...}\left[ \sum^\infty_{t=0}\gamma^tr(s_t,a_t)\right]+\\
        +\alpha\max(0,\mathbb{E}_{a\sim \pi_{\theta_{old}}}\left[D_{KL}(\pi_{\theta_{old}(a)},\pi_{\theta_{new}}(a))\right]-\delta_{TR})  
    \end{split}
\end{gather}

For matters of simplicity of implementation, Schulman et al. (2015) \cite{Schulman2015} rewrote equation \eqref{eq:pol_loss_paper} in terms of collected experience (cf. equation \eqref{eq:pol_loss_paper_with_GAE}), as a function of two different stochastic policies, $\pi_{\theta_{new}}$ and $\pi_{\theta_{old}}$, and the $GAE$.

\begin{equation}\label{eq:pol_loss_paper_with_GAE}
    \begin{split}
        L_P = -\mathbb{E}_{s}\mathbb{E}_{a\sim \pi_{\theta_{old}}}\left[GAE_{\theta_{old}}(s)\frac{\pi_{\theta_{new}}(a)}{\pi_{\theta_{old}}(a)}\right]+\\
        +\alpha\max(0,\mathbb{E}_{a\sim \pi_{\theta_{old}}}\left[D_{KL}(\pi_{\theta_{old}}(a),\pi_{\theta_{new}}(a))\right]-\delta_{TR})
    \end{split}
\end{equation}

Both versions of TRPO use the same exploration strategy: the output of the Policy NN is used as the mean of a multivariate normal distribution whose covariance matrix is part of the policy trainable parameters.

\subsection{Hindsight Experience Replay}\label{sec:her}

The key idea of HER (Andrychowicz et al. 2018 \cite{Andrychowicz2018}) is to store in the replay buffers not only the experience collected from interacting with the environment, but also about experiences that would be obtained, had the agent followed a different goal. HER was proposed as a complement of sparse rewards.

\subsection{Prioritized Experience Replay}\label{sec:per}

When working with replay buffers, randomly sampling experience can be outperformed by a more sophisticated heuristic. Schaul et al. (2016) \cite{Schaul2016a} proposed Prioritized Experience Replay, sampling experience according to their significance, measured by the TD-error. Besides, the strict required performance (cf. table \ref{tab:objectives}) causes the agent to seldom achieve success and the training dataset to be imbalanced. Thus, the agent simply overfits the "failure" class. Narasimhan et al. (2015) \cite{Narasimhan2015} have addressed this problem by forcing 25\% of the training dataset to be sampled from the less represented class.

\section{Application of RL to the Missile’s Flight}
\label{sec:imple}

One episode is defined as the attempt of following a 5s-long $a_z$ reference signal consisting of two consecutive steps whose amplitude and rise times are randomly generated except when deploying the agent for testing purposes (Cortez et al. 2020 \cite{Cortez2020}). With a sampling time of 1ms, an episode includes 5000 steps, from which 2400 are defined as transition periods (the 600 steps composing 0.6s after each of the four rise times), whilst the remaining 2600 are resting periods.

From the requirements defined in section \ref{section:requirements}, the RL problem was formulated as the training of an agent that succeeds when the performance of an episode meets the levels defined in table \ref{tab:objectives} in terms of the
maximum stationary tracking error $|e_z|_{\max,r}$, of the overshoot, of the actuation magnitude $|\eta|_{\max}$ and of the actuation noise levels both in resting ($\eta_{noise,r}$) and transition ($\eta_{noise,t}$) periods.

\begin{table}[!htb]
\caption{Performance Objectives}
  \begin{tabular}{lllllccc}
    \hline
    \textbf{Requirement} &  \textbf{Achieved Value} & \\
    \hline
    $|e_z|_{\max,r}$   &   0.5    &  [g] \\
Overshoot   &   20   &   [\%] \\
$|\eta|_{\max}$   &   15  &   [º] \\
$\eta_{noise,r}$   &   1    &   [rad] \\
$\eta_{noise,t}$   &   0.2    &   [rad] \\
    \hline
  \end{tabular}
  \label{tab:objectives}
\end{table}

\subsection{Algorithm}
\label{section:implemented_algorithm}

As explained in section \ref{sec:intro}, the current problem required an on-policy model-free RL algorithm. Among them, not only is TRPO a current state-of-the-art algorithm (cf. section \ref{sec:topic_overview}), but it also presents the attractiveness of the trust region search, avoiding sudden drops during the training progress, which is a very interesting feature to be explored by the industry, whose mindset is often aiming at robust results. TRPO was, therefore, the most suitable choice.

\subsubsection{Modifications to original TRPO}
\label{section:modifications2trpo}

The present implementation was inspired in the implementations proposed by Schulman et al. (2015) \cite{Schulman2015} and by Kangin et al. (2019) \cite{Kangin2019} (cf. section \ref{sec:trpo}). There are several differences, though:

\begin{enumerate}
    \item The reward function is given by equation \eqref{eq:orf}, whose relative weights $w_i$ can be found in Cortez et al. (2020) \cite{Cortez2020}.

    \begin{gather}
        f_1 = -w_1.|e_z|\\
        f_2 = \left\{
            \begin{array}{ll}
                  0 & \mbox{if } |\eta|<\eta_{max} \\
                  -w_2 & \mbox{otherwise }
            \end{array} 
        \right.\label{eq:rew_term_eta_penalty} \\
        \eta_{slope} = \frac{\Delta\eta}{t_s}\\
        f_3 = -w_3.|\eta_{slope}|   \label{eq:rew_term_eta_slope}   \\
        \begin{split}
            condition = |e_z|<3g \wedge |\eta|<0.2 \wedge  \\
            \wedge |e_u|<e_{u,\max}
        \end{split} \\
        f_4 = \left\{
            \begin{array}{ll}
                  w_4.\frac{e_{u,\max} - |e_u|}{e_{u,\max}} & \mbox{if } condition\\
                  0 & \mbox{otherwise }
            \end{array} 
        \right. \label{eq:rew_term_bonus}   \\
        r = \sum_{i=1}^4f_i   \label{eq:orf}
    \end{gather}
    
    \item Both neural networks have got three hidden layers whose sizes - $h_1$, $h_2$ and $h_3$ - are related with the observations vector (their input size) and with the actions vector (output size of the Policy NN).
    
    \item Observations are normalized (cf. equation \eqref{eq:scaler}) so that the learning process can cope with the different domains of each feature.
    
    \begin{equation}
        obs_{norm} = \frac{obs-\mu_{online}}{\sigma_{online}}
        \label{eq:scaler}
    \end{equation}
    
    In equation \eqref{eq:scaler}, $\mu_{online}$ and $\sigma_{online}$ are the running mean value and the running variance of the set of observations collected along the whole training process, which are updated with information about the newly collected observations after each training episode.
    
    \item The proposed exploration strategy is deeply rooted on Kangin et al.'s (2019) \cite{Kangin2019}, meaning that, the new action $\eta$ is sampled from a normal distribution (cf. equation \eqref{eq:sample_eta}) whose mean is the output of the policy neural network and whose variance is obtained according to equation \eqref{eq:eta_variance}.

    \begin{gather}
        \eta \sim N(\mu_{\eta},\sigma^2_{\eta}) \label{eq:sample_eta}\\
        \sigma^2_{\eta} = e^{\sigma^2_{\log}} \label{eq:eta_variance}
    \end{gather}
    
    Although similar, this strategy differs from the original in $\sigma^2_{\log}$ (cf. equation \eqref{eq:log_var}), which is directly influenced by the tracking error $e_z$ through $\sigma^2_{\log, tune}$.

    \begin{equation}
        \sigma^2_{\log} = \sigma^2_{\log, train}+\sigma^2_{\log, tune}(e_z)    \label{eq:log_var}
    \end{equation}
    
    \item Equation \eqref{eq:policy_loss_with_GAE} was used as the loss function of the policy parameters, modifying equation \eqref{eq:pol_loss_paper_with_GAE} in order to emphasize the need of reducing $D_{KL}$ with a term that linearly penalizes $D_{KL}$ and a quadratic term that aims at fine-tuning it so that it is closer to the trust region radius $\delta_{TR}$, encouraging as big update steps as possible.

    \begin{gather}
        L_1 = \mathbb{E}_{s}\mathbb{E}_{\eta \sim \pi_{\theta_{old}}}\left[GAE_{\pi_{\theta_{old}}}(s)\frac{\pi_{\theta_{new}}(\eta)}{\pi_{\theta_{old}}(\eta)}\right]    \label{eq:policy_loss_with_GAE_1st}    \\
        L_2 = \mathbb{E}_{\eta\sim \pi_{\theta_{old}}}\left[D_{KL}(\pi_{\theta_{old}}(\eta),\pi_{\theta_{new}}(\eta))\right]   \\
        \begin{split}
            L_P = -L_1+\alpha\left(\max(0,L_2-\delta_{TR})\right)^2+\beta L_2
        \end{split} \label{eq:policy_loss_with_GAE}
    \end{gather}
    
    Having the previously mentioned exploration strategy, $\pi_{\theta}$ is a Gaussian distribution over the continuous action space (cf. equation \eqref{eq:pi}, where $\mu_{\eta}$ and $\theta_{\eta}$ are defined in equation \eqref{eq:sample_eta}).
    
    \begin{equation}
        \pi_{\theta}(\eta) = \frac{1}{\sigma_{\eta}\sqrt{2\pi}}e^{-\frac{\left(\eta-\mu_{\eta}\right)^2}{2\sigma_{\eta}^2}}  \label{eq:pi}
    \end{equation}
    
    Hence, $L_1$ (cf. equation \eqref{eq:policy_loss_with_GAE_1st}) is given by equation \eqref{eq:pis_GAE_over_batch}, where $n$ is the number of samples in the training batch, assuming that all samples in the training batch are independent and identically distributed\footnote{We can assume they are (i) independent because they are sampled from the replay buffer (stage \ref{list:algo_bper} of algorithm \ref{algorithm:trpo_implemented}, section \ref{section:algorithm_description}), breaking the causality correlation that the temporal sequence could entail, and (ii) identically distributed because the exploration strategy is always the same and, therefore, the stochastic policy $\pi_{\theta}$ is always a Gaussian distribution over the action space (cf. equation \eqref{eq:pi}).}.
    
    \begin{equation}
        L_1 = \left[\frac{1}{n}\sum^n_{i=1}GAE_{\pi_{\theta_{old}}}s_i\right].\Pi_{i=1}^n\left[\frac{\pi_{\theta_{new}}(\eta_i)}{\pi_{\theta_{old}}(\eta_i)}\right]   \label{eq:pis_GAE_over_batch} 
    \end{equation}
    
    Moreover, $L_2$ is given by equation \eqref{eq:expected_D_KL}.
    
    \begin{equation}
        L_2 = \frac{1}{n}\sum_{i=1}^nD_{KL}(\pi_{\theta_{old}}(\eta_i),\pi_{\theta_{new}}(\eta_i))    \label{eq:expected_D_KL}
    \end{equation}
    
    \item ADAM was used as the optimizer of both NN for its wide cross-range success and acceptability as the default optimizer of most ML applications.
\end{enumerate}

\subsubsection{Hindsight Experience Replay}
\label{section:her}
The present goal is not defined by achieving a certain final state as Andrychowicz et al. (2018) \cite{Andrychowicz2018} proposed, but, instead, a certain performance in the whole sequence of states that constitutes an episode. For this reason, choosing a different goal must mean, in this case, to follow a different reference signal. After collecting a full episode, those trajectories are replayed with new goals, which are sampled according to two different strategies. These strategies dictate the choice of the amplitudes of the two consecutive steps of each new reference signal.

The first strategy - \textit{mean} strategy - consists of choosing the amplitudes of the steps of the command signal as the mean values of the measured acceleration during the first and second resting periods, respectively. Similarly, the second strategy - \textit{final} strategy - consists of choosing them as the last values of the measured signal during each resting period. Apart from the step amplitudes, all the other original parameters (Cortez et al. 2020 \cite{Cortez2020}) of the reference signal are kept.

\subsubsection{Balanced Prioritized Experience Replay}
\label{section:prioritized_exp_replay}

Being $l_i$ the priority level of the experience collected in step $i$ (cf. equation \eqref{eq:priority}, with $e_{u,\max}=0.01$), $N_j$ the number of steps with priority level $j$ and $\rho_j$ the proportion of steps with priority level $j$ desired in the training datasets, BPER was implemented according to algorithm \ref{algorithm:prio_exp_replay}.

\begin{equation}\label{eq:priority}
    l_i = \left\{
        \begin{array}{ll}
              1 & \mbox{if } |e_z|_i<0.5g \wedge |\eta|_i<\frac{|\eta|_{max}}{2} \wedge  |e_u|_i<e_{u,\max} \\
              0 & \mbox{otherwise }
        \end{array} 
    \right.
\end{equation}

Notice that $P(i)$ and $p_i$ follow the notation of Schaul et al. (2016) \cite{Schaul2016a} (assuming $\alpha=1$), in which $p_i = \frac{1}{rank(i)}$ matches the rank-based prioritization with $rank(i)$ meaning the ordinal position of step $i$ when all steps in the replay buffers are ordered by the magnitude of their temporal differences. 

\begin{algorithm}\label{algorithm:prio_exp_replay}
\SetAlgoLined
    \eIf{$N_1<0.25(N_0+N_1)$}{
        $\rho_1 = 0.25$\\
        $s_j = \sum_i p_i, \forall_i (l_i=j)$ with $j\in\{0,1\}$\\}{
        $\rho_1 = 0.5$\\
        $s_0 = s_1 = \sum_i p_i, \forall_i$\\}
    $\rho_0 = 1-\rho_1$\\
    $P(i) = \left\{
        \begin{array}{ll}
              \rho_1\times\frac{p_i}{s_1} & \mbox{if } l_i=1\\
              \rho_0\times\frac{p_i}{s_0} & \mbox{otherwise }
        \end{array} 
    \right.$ \\
    \caption{Balanced Prioritized Experience Replay}
\end{algorithm}

As condensed in algorithm \ref{algorithm:prio_exp_replay}, when there is less than 25\% of successful steps in the replay buffers, the successful and unsuccessful subsets of the replay buffers are sampled separately, with 25\% coming from the successful subset. In a posterior phase of training, when there is already more than 25\% of successful steps, both subsets are molten. In either cases, sampling is always done according to the temporal differences, i.e., a step with higher temporal difference has got a higher chance of being sampled.

\subsubsection{Scheduled Experience Replay}
\label{section:SER}
Having HER (cf. section \ref{section:her}) and BPER (cf. section \ref{section:prioritized_exp_replay}) dependent on a condition - the SER condition - is hereby defined as Scheduled Experience Replay (SER). The SER condition is exemplified in equation \eqref{eq:SER_condition}, where $\Bar{e}_{z,past}$ stands for the mean tracking error of the previously collected episode .

\begin{equation}
    \Bar{e}_{z, past}\leq2g
    \label{eq:SER_condition}
\end{equation}

Contrarily to its original context (cf. section \ref{sec:her}), the reward function is not sparse and was already able of achieving near-target performance without HER. The hypothesis, in this case, is that HER can be a complementary feature, by activating it only when the agent converges to suboptimal policies.

Moreover, without HER, BPER adds less benefit, since there is no special reason for the agent to believe that some part of the collected experience is more significant than other.

\subsubsection{Algorithm Description}\label{section:algorithm_description}

\begin{algorithm}\label{algorithm:trpo_implemented}
\SetAlgoLined
    Initialization \\
    \While{training}{
    
        \begin{enumerate}
            \item Collect experience, sampling actions from policy $\pi$ \label{list:algo_collect_batch}\\
            \item Augment the collected experience with synthetic successful episodes (SER) \label{list:algo_her}\\
            \item $(a_t,s_t,r_t) \leftarrow V'(s_t)$ and $GAE(s_t)$ \textbf{for all} $(a_t,s_t,r_t)$ \textbf{in} $T$, \textbf{for all} $T$ \textbf{in} $B$   \label{list:algo_add2traj}\\
            \item Store the newly collected experience in the replay buffer     \label{list:algo_put_in_rb}\\
            \item Sample the training sets from the replay buffer (SER)    \label{list:algo_bper}\\
            \item Update all value and policy parameters \label{list:algo_update_nn}\\
            \item Update trust region  \label{list:algo_update_tr}\\
        \end{enumerate}
    }
    \caption{Implemented TRPO with a Replay Buffer and SER}
\end{algorithm}

\begin{enumerate}
    \item One batch $B$ of trajectories $T$, is collected.
    
    \item If the SER condition (cf. section \ref{section:SER}) holds, $B$ is augmented according to HER (cf. section \ref{section:her}).
    
    \item The targets for the Value NN $V'(s_t)$ and the $GAE(s_t)$ are computed and added to the trajectories.
    
    \item $B$ is stored in the replay buffer, discarding the oldest batch: the replay buffer $R$ contains data collected from the last policies and works as a FIFO queue.
    
    \item If the SER condition (cf. section \ref{section:SER}) holds, the training dataset is sampled from $R$ according to BPER (cf. section \ref{section:prioritized_exp_replay}). Otherwise, the entire information available in $R$ is used as training dataset.
    
    \item The value parameters and the policy parameters are updated.
    
    \item The trust region parameters are updated.
\end{enumerate}

\subsection{Methodology}   \label{section:methodology}
As further detailed in \cite{Cortez2020}, the established methodology (i) progressively increases the amplitude of the randomly generated command signal and (ii) intermediate testing of the agent's performance against a  -10g/10g double step without exploration, in order , respectively, (i) to avoid overfitting and (ii) to decide whether or not to finish the training process.

\subsection{Robustness Assessments}
\label{section:robustness_implementation}

The formal mathematical guarantee of  robustness of a RL agent composed of neural networks cannot be done in the same terms as the one of linear controllers. It was, hence, evaluated by deploying the nominal agent in non-nominal environments. Apart from testing its performance, the hypothesis is also that training this agent in the latter can improve its robustness. To do so, the training of the best found nominal agent was resumed in the presence of non-nominalities (cf. section \ref{section:non-nominal_env_implementation}). This training and its resulting best found agent are henceforward called \textit{robustifying training} and \textit{robustified agent}, respectively.

\subsubsection{Robustifying  Trainings}
\label{section:non-nominal_env_implementation}
Three different modifications were separately made in the provided model (cf. section \ref{sec:model}), in order to obtain three different non-nominal environments, each of them modelling latency, estimation uncertainty in the nominal estimated Mach number $M_{nom}$ and height $h_{nom}$ (cf. equations \eqref{eq:estim_uncert_M} and \eqref{eq:estim_uncert_h}) and parametric uncertainty in the nominal aerodynamic coefficients $C_{z,nom}$ and $C_{m,nom}$ (cf. equations \eqref{eq:param_uncert_cfbb_z} and \eqref{eq:param_uncert_cmbb_m}).

\begin{gather}
    M = (1+\Delta M)\times M_{nom}\label{eq:estim_uncert_M}   \\
    h = (1+\Delta h)\times h_{nom}\label{eq:estim_uncert_h}    \\
    C_z = (1+\Delta C_z)\times C_{z,nom}\label{eq:param_uncert_cfbb_z}   \\
    C_m = (1+\Delta C_m)\times C_{m,nom}\label{eq:param_uncert_cmbb_m}
\end{gather}

In the case of non-nominal environments including latency, the range of possible values is $[0,l_{max}]\cap\mathbb{N}^0$, whereas in the other cases, the uncertainty is assumed to be normally distributed, meaning that, following Peter's (2018) \cite{Peter2018} line of thought, its domain is $[\mu-3\sigma, \mu+3\sigma]$\footnote{To be accurate, this interval covers only 99.73\% of the possible values, but it is assumed to be the whole spectrum of possible values.}.

Before each new episode of a robustifying training, the non-nominality new value (either latency or one of the uncertainties) was sampled from a uniform distribution over its domain and kept constant during the entire episode.
In each case, four different values were tried for the bounds of the range of possibilities (either $l_{max}$ or $3\sigma$):

\begin{enumerate}
    \item $l_{max} \in \{1, 3, 5,  10\}$ [ms]
    \item $(3\sigma_{estimation}) \in \{1,2,3,5\}$ [\%]
    \item $(3\sigma_{parametric}) \in \{5,7,10,15\}$ [\%]
\end{enumerate}

Values whose robustying training had diverged after 2500 episodes were discarded. The remaining were run for a total of 5000 episodes (cf. section \ref{section:robustness}).

\section{Results}
\label{sec:resul}

\subsection{Expected Results}\label{section:expec_res}

Empirically, it has been seen that agents that struggle to control $\eta$'s magnitude within its bounds (cf. table \ref{tab:objectives}) are unable of achieving low error levels without increasing the level of noise, if they can do it at all. In other words, the only way of having good tracking results with unbounded actions is to fall into a bang-bang control-like situation. Therefore, the best found agent will have to start by learning to use only bounded and smooth action values, which will allow it to, then, start decreasing the tracking error.

Notice that, while the tracking error and the noise measures are expected to tend to 0, $\eta$'s magnitude is not, since it would mean that the agent had given up actuating in the environment. Hence, it is expected that, at some point in training, the latter stabilizes.

Moreover, the exploration strategy proposed in section \ref{section:modifications2trpo} insert some variance in the output of the policy neural network, meaning that the noise measures are also not expected to reach exactly 0.

\subsection{Best Found Agent}
\label{section:optimal_agent}

\begin{figure}[!htb]
    \begin{subfigmatrix}{2}
        \subfigure[Tracking performance]{\includegraphics[width=0.49\linewidth]{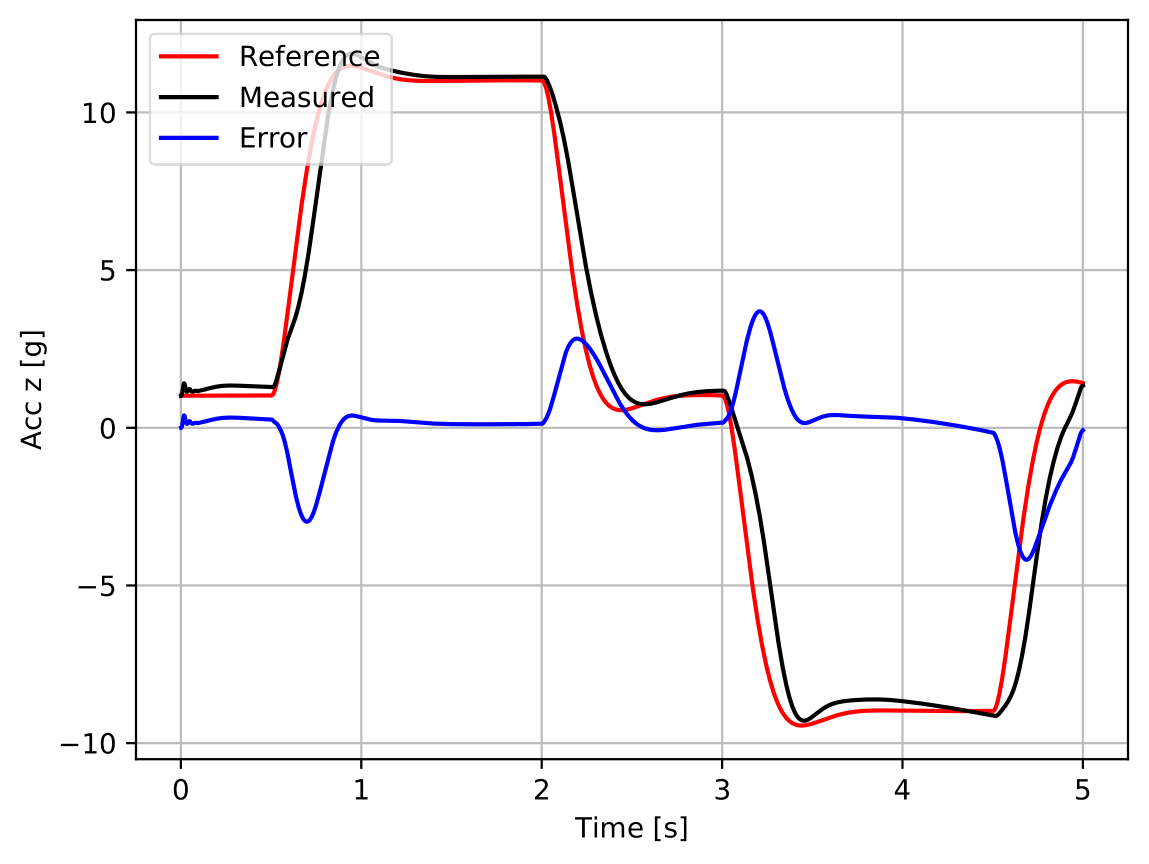}}
        \subfigure[Action, $\eta$]{\includegraphics[width=0.49\linewidth]{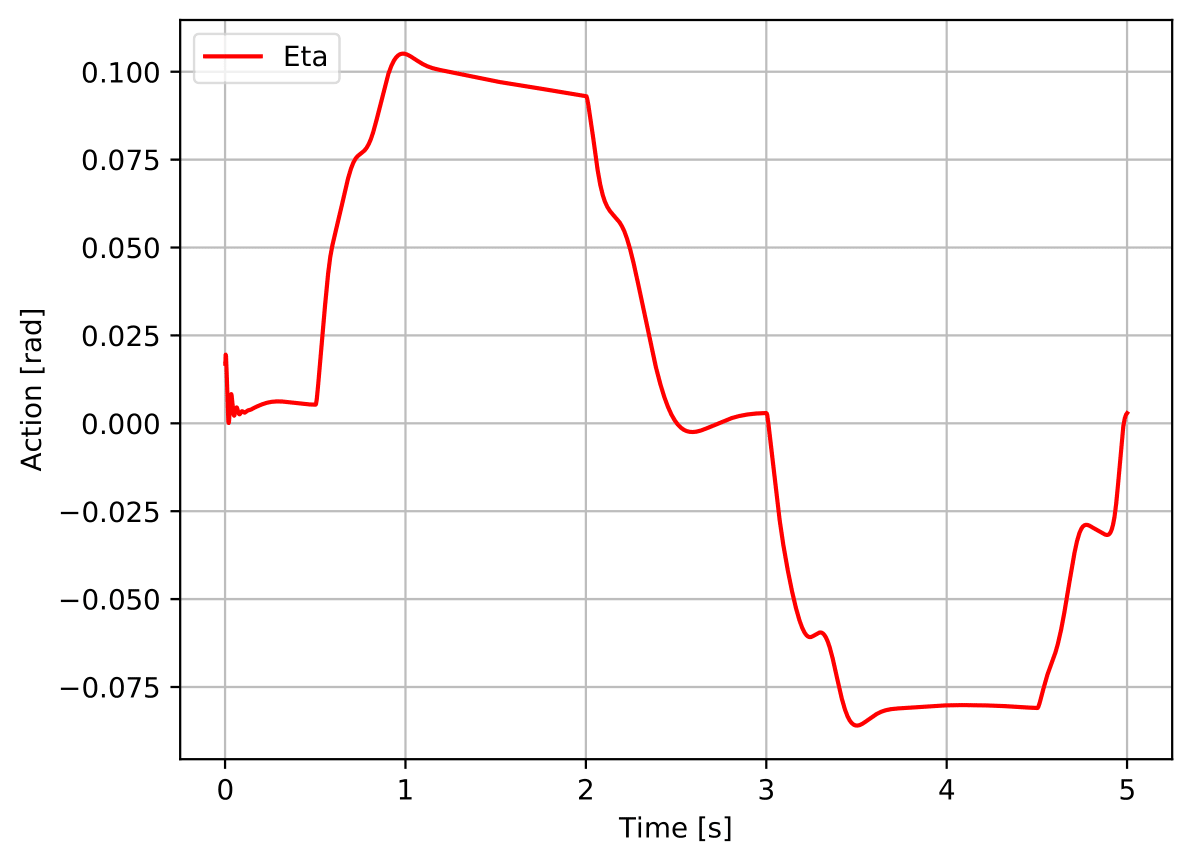}}
    \end{subfigmatrix}
    \caption[Test performance of the Best Found Agent]{Test performance of the Best Found Agent}
    \label{fig:opt_test}
\end{figure}

As figure \ref{fig:opt_test} evidences, the agent is clearly able of controlling the measured acceleration and to track the reference signal it is fed with, satisfying all the performance requirements defined in table \ref{tab:objectives} (cf. table table \ref{tab:opt_test}).

\begin{table}[!htb]
\caption{Performance achieved by the Best Found Nominal Agent}
  \begin{tabular}{lllllccc}
    \hline
    \textbf{Requirement} &  \textbf{Achieved Value} & \\
    \hline
    $|e_z|_{\max,r}$   &   0.4214    &  [g] \\
Overshoot   &   8.480   &   [\%] \\
$|\eta|_{\max}$   &   0.1052  &   [rad] \\
$\eta_{noise,r}$   &   0.04222    &   [rad] \\
$\eta_{noise,t}$   &   0.005513    &   [rad] \\
    \hline
  \end{tabular}
  \label{tab:opt_test}
\end{table}

\subsection{Reproducibility Assessments}
\label{section:reproducibility_resul}

\begin{figure}[!htb]
    \begin{subfigmatrix}{2}
        \subfigure[Tracking performance]{\includegraphics[width=0.49\linewidth]{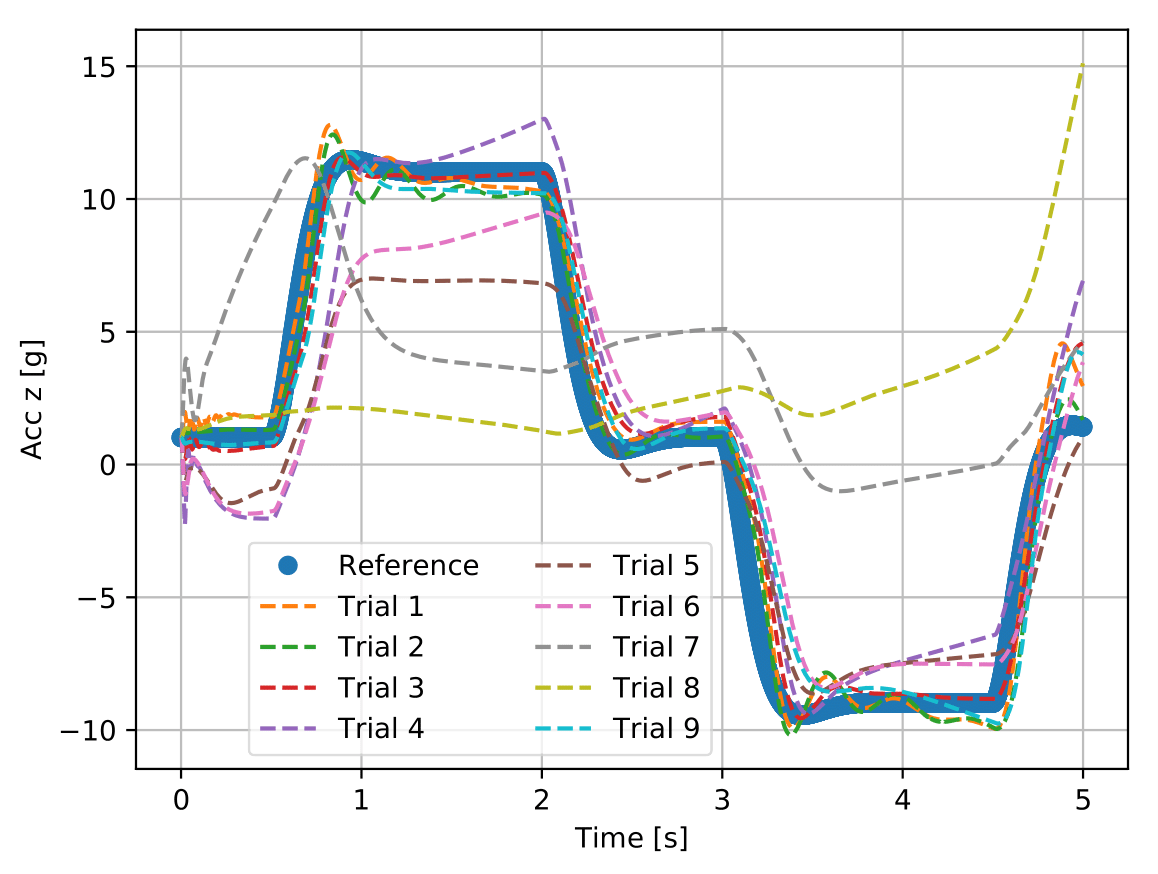}}
        \subfigure[Action, $\eta$]{\includegraphics[width=0.49\linewidth]{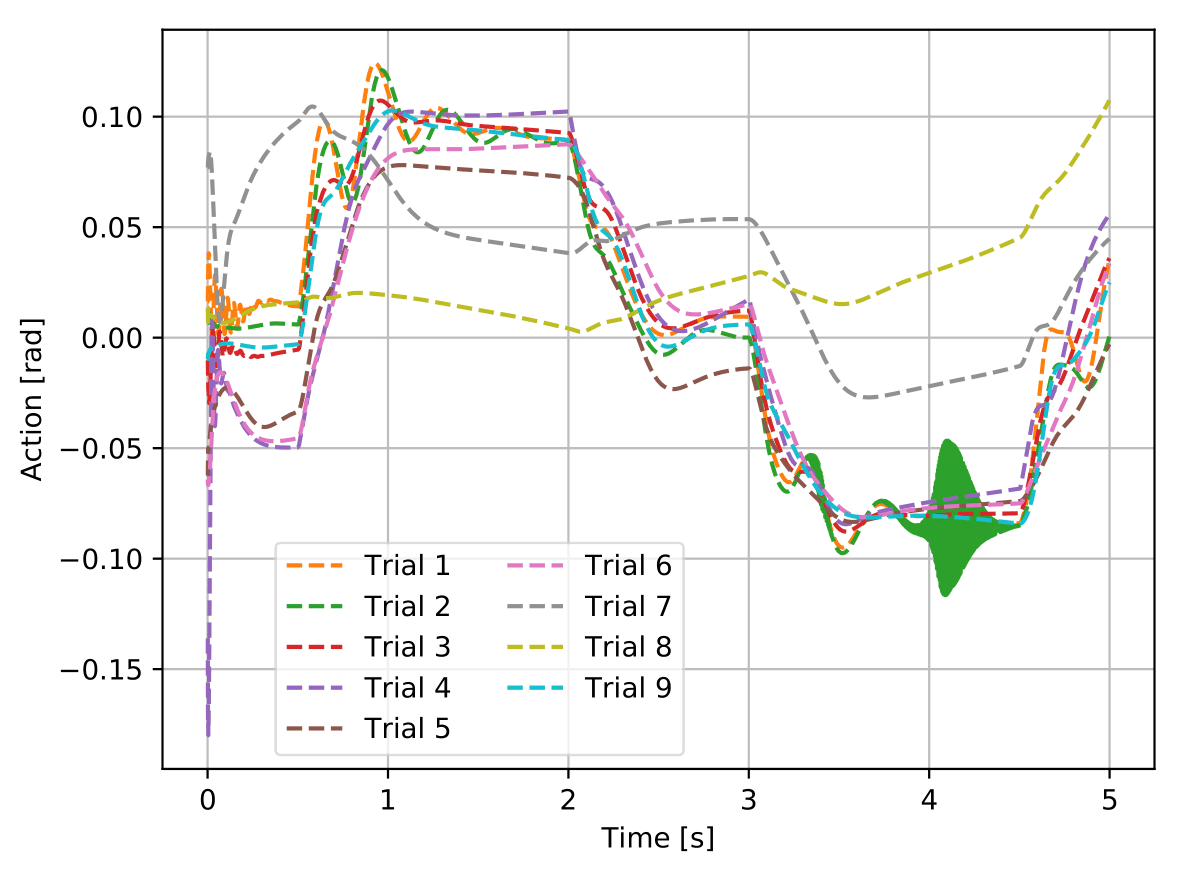}}
    \end{subfigmatrix}
    \caption{Nominal test performance of the Reproducibility Trials}
    \label{fig:repro_test_trials}
\end{figure}

Figure \ref{fig:repro_test_trials} shows the tests of the best agents obtained during each of the nine trainings run to assess the reproducibility of the best found nominal agent (cf. section \ref{section:optimal_agent}) and it is possible to see that none of them achieved the target performance, i.e., none meets all the performance criteria established in table \ref{tab:objectives}. Although most of the trials can be considered far from the initial random policy, it is possible to verify that trials 7 and 8 present a very poor tracking performance and that trial 2's action signal is not smooth.

\subsection{Robustness Assessments}
\label{section:robustness}

\subsubsection{Latency}\label{section:robustness_latency_results}

\begin{figure}[!htb]
  \centering
  \includegraphics[width=1\linewidth]{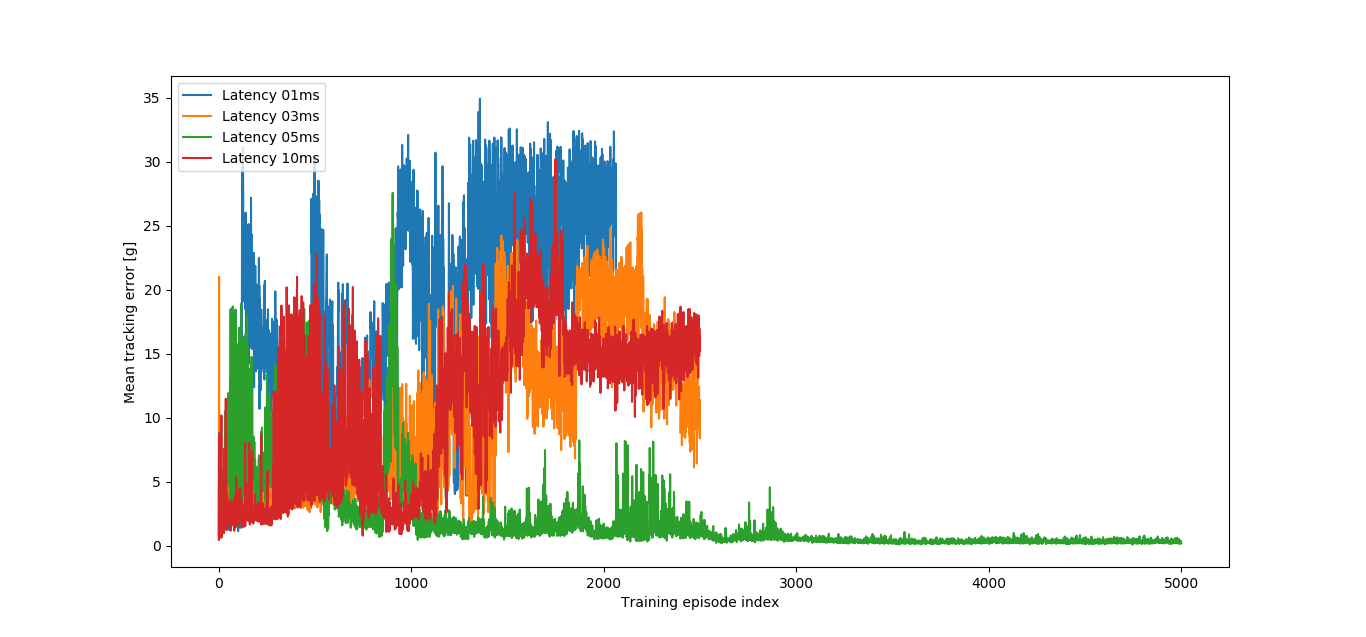}
  \caption[Mean tracking error of latency robustifying trainings]{Mean tracking error of latency robustifying trainings}
  \label{fig:rob_latency_trains}
\end{figure}

\begin{figure}[!htb]
    \begin{subfigmatrix}{2}
        \subfigure[Tracking performance]{\includegraphics[width=0.49\linewidth]{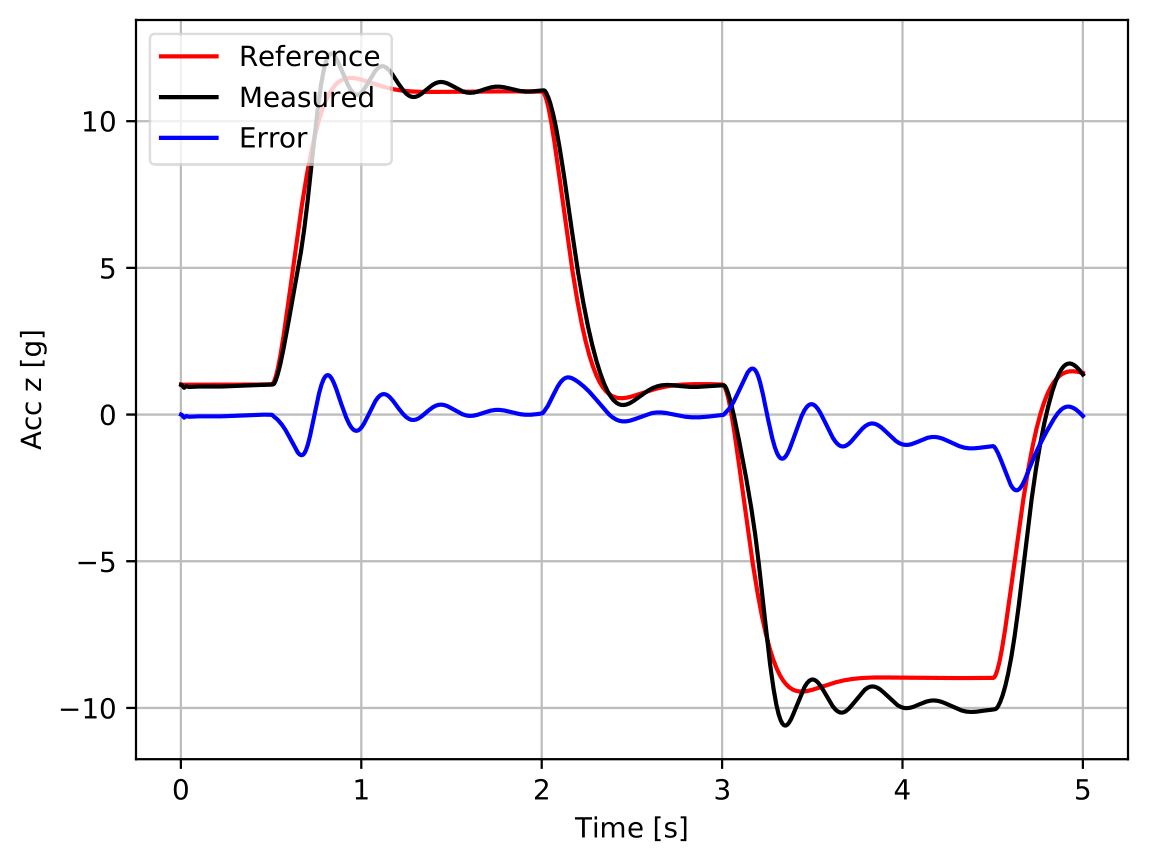}}
        \subfigure[Action, $\eta$]{\includegraphics[width=0.49\linewidth]{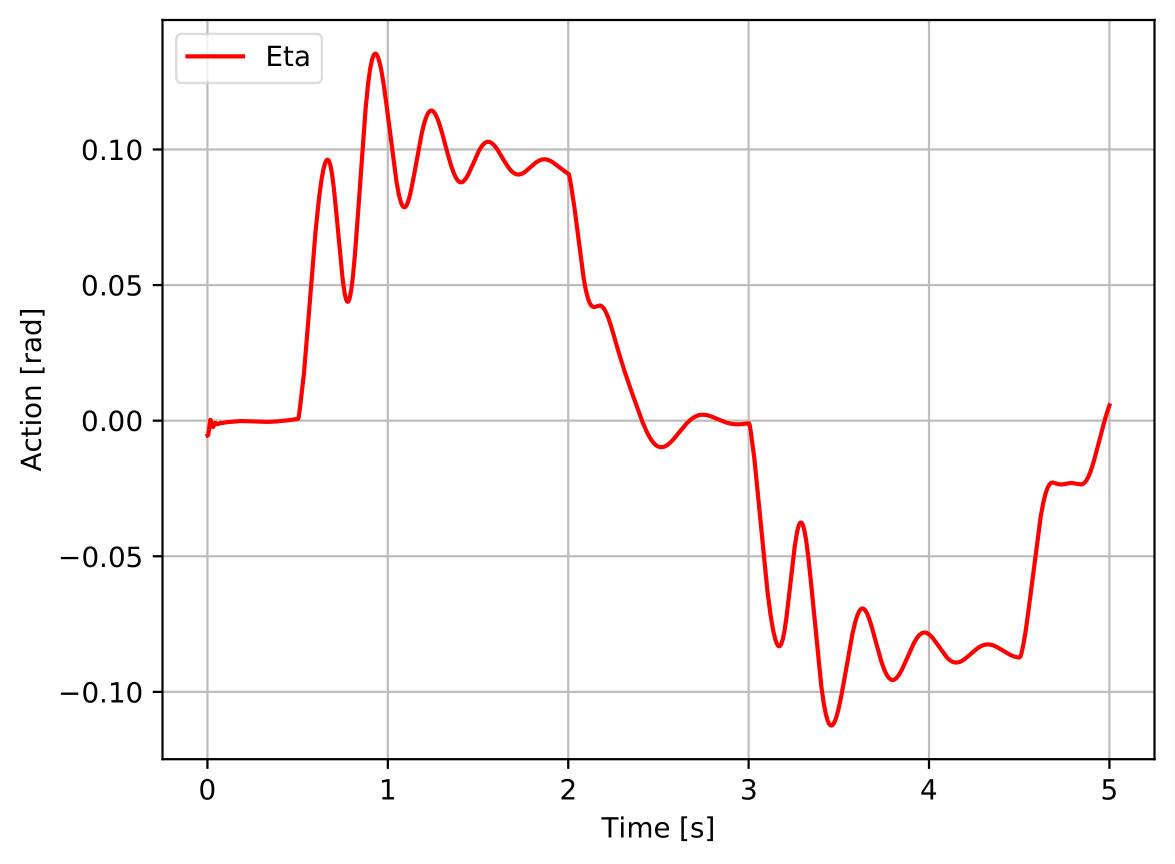}}
    \end{subfigmatrix}
    \caption[Nominal Performance of the Latency Robustified Agent]{Nominal Performance of the Latency Robustified Agent}
    \label{fig:nom_test_lat_rob_agent}
\end{figure}

As figure \ref{fig:rob_latency_trains} shows, from the four different values of $l_{max}$, only 5ms converged after 2500 episodes, meaning that it was the only robustifying training being run for a total of 5000 episodes, during which the best agent found was defined as the Latency Robustified Agent. The nominal performance (cf. figure \ref{fig:nom_test_lat_rob_agent}) is damaged, having a less stable action signal and a poorer tracking performance.

Furthermore, its performance in environments with latency (0ms to 40ms) did not provide any enhancement, as its success rate is low in all quantities of interest (cf. table \ref{tab:lat_test_comparison}).

\begin{table}[!htb]
  \renewcommand{\arraystretch}{1.2} 
\caption{Robustified Agent success rate in improving the robustness to Latency of the Nominal Agent}
  \begin{tabular}{lcc}
    \hline
    \textbf{Requirement} &  \textbf{Success \%} \\
    \hline
    $|e_z|_{\max,r}$ &  0.00    \\
    Overshoot   &   25.00  \\
    $\eta_{\max}$  &   0.00   \\
    $\eta_{noise,r}$    &    7.50   \\
    $\eta_{noise,t}$    &    5.00   \\
    \hline
  \end{tabular}
  \label{tab:lat_test_comparison}
\end{table}

Since the Latency Robustified Agent was worse than the Nominal Agent in both nominal and non-nominal environments, it lost in both Performance and Robustness categories. Thus, it is possible to say, in general terms, that, concerning latency, the robustifying trainings failed.

\subsection{Estimation Uncertainty}\label{section:robustness_estimation_uncertainty}

\begin{figure}[!htb]
  \centering
  \includegraphics[width=1\linewidth]{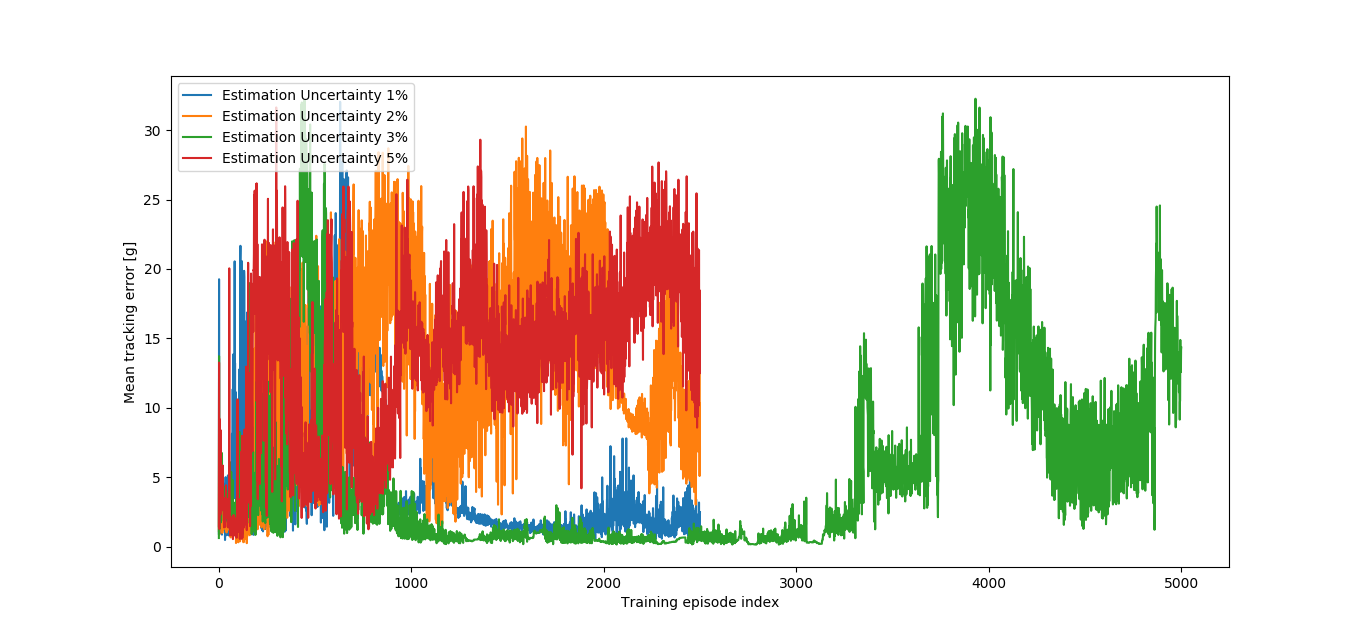}
  \caption[Mean tracking error of estimation uncertainty robustifying trainings]{Mean tracking error of estimation uncertainty robustifying trainings}
  \label{fig:rob_estim_trains}
\end{figure}

As figure \ref{fig:rob_estim_trains} shows, from the four different values of $3\sigma$ tried, only the 3\% one converged after 2500 episodes, meaning that it was the only robustifying training being run for a total of 5000 episodes, during which the best agent found was defined as the Estimation Uncertainty Robustified Agent. The nominal performance is improved, having a lower overshoot and a smoother action signal.

\begin{figure}[!htb]
    \begin{subfigmatrix}{2}
        \subfigure[Tracking performance]{\includegraphics[width=0.49\linewidth]{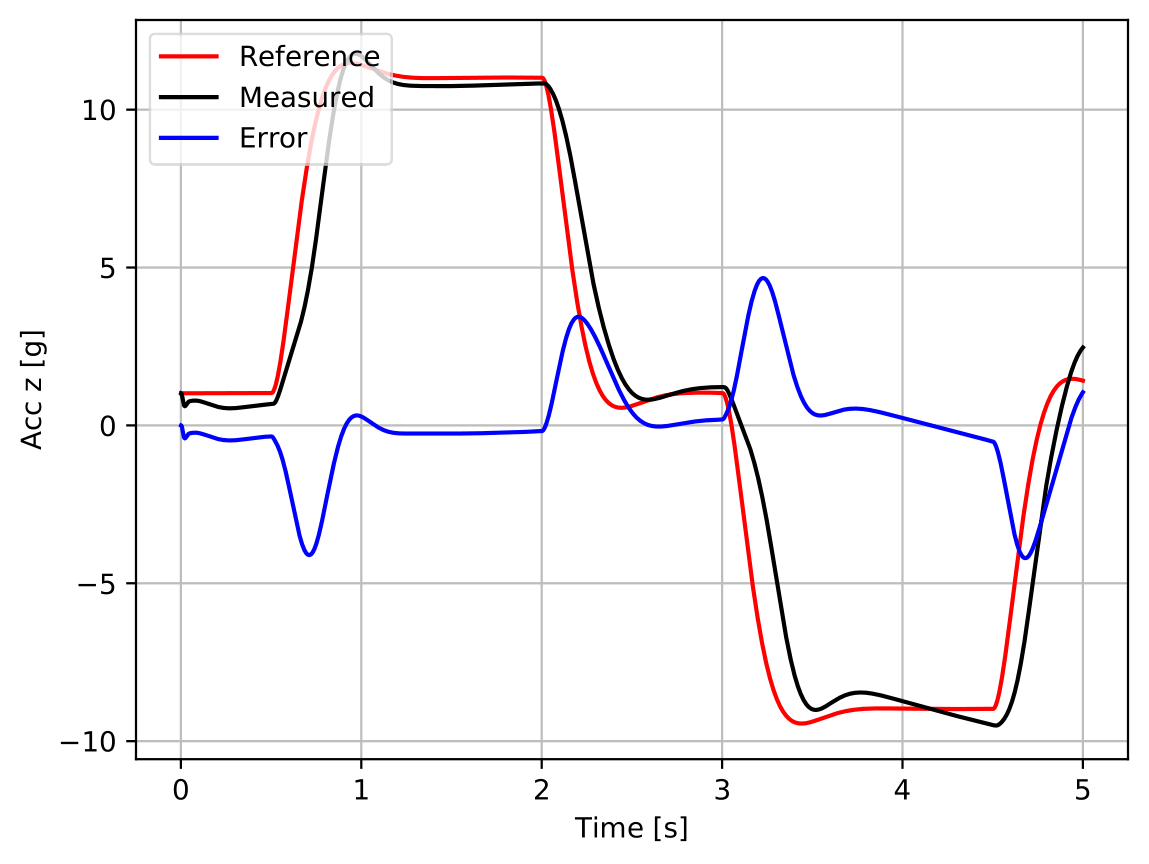}}
        \subfigure[Action, $\eta$]{\includegraphics[width=0.49\linewidth]{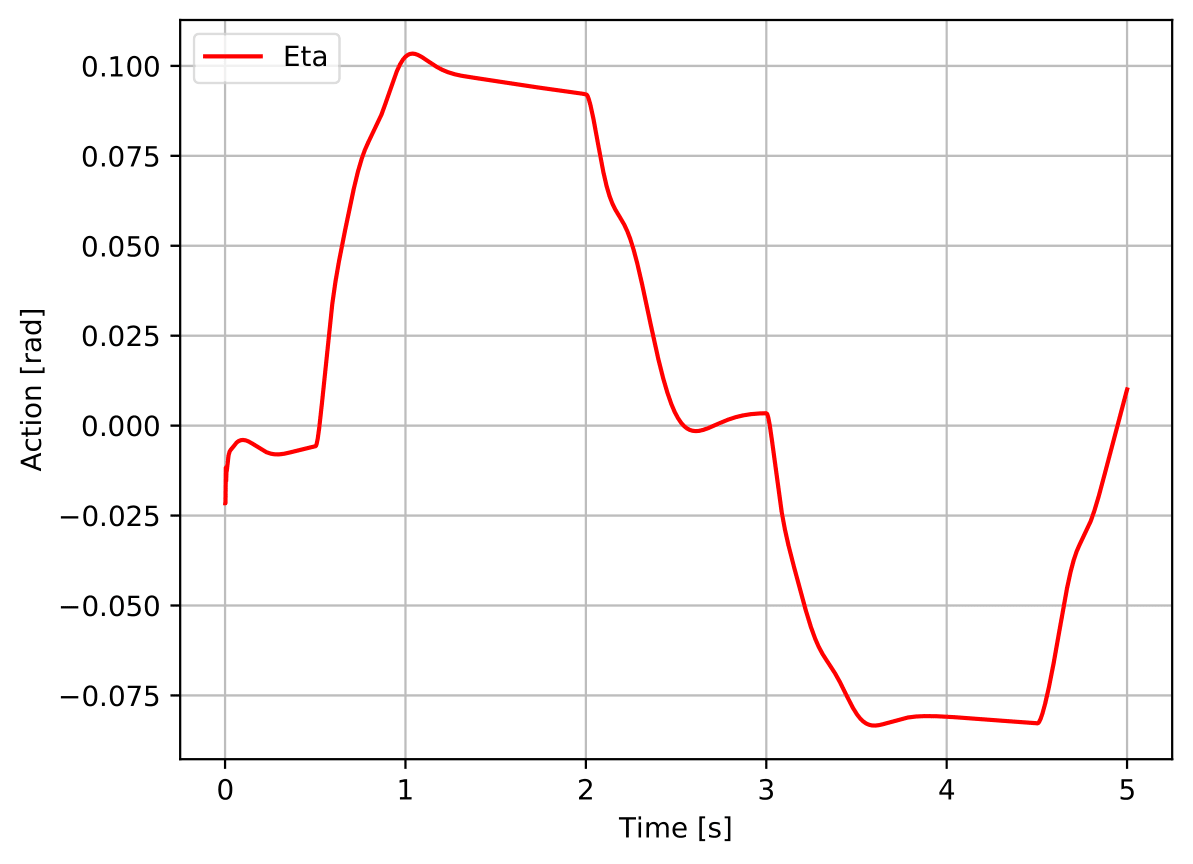}}
    \end{subfigmatrix}
    \caption[Nominal Performance of the Estimation Uncertainty Robustified Agent]{Nominal Performance of the Estimation Uncertainty Robustified Agent}
    \label{fig:nom_test_estim_rob_agent}
\end{figure}

Furthermore, it enhanced the performance of the Nominal Agent in environments with estimation uncertainty (-10\% to 10\%), having achieved high success rates (cf. table \ref{tab:estim_test_comparison}).

\begin{table}[!htb]
  \renewcommand{\arraystretch}{1.2} 
\caption{Robustified Agent success rate in improving the robustness to Estimation Uncertainty of the Nominal Agent}
  \begin{tabular}{lcc}
    \hline
    \textbf{Requirement} &  \textbf{Success \%} \\
    \hline
    $|e_z|_{\max,r}$    & 60.38  \\
    Overshoot   &   66.39\\
    $\eta_{\max}$  &    91.97\\
    $\eta_{noise,r}$    &    77.51\\
    $\eta_{noise,t}$    &    82.33\\
    \hline
  \end{tabular}
  \label{tab:estim_test_comparison}
\end{table}

Since the Estimation Uncertainty Robustified Agent was better than the Nominal Agent in both nominal and non-nominal environments, it won in both Performance and Robustness categories. Thus, it is possible to say, in general terms, that, concerning estimation uncertainty, the robustifying trainings succeeded.

\subsection{Parametric Uncertainty}\label{section:robustness_parametric_uncertainty}

\begin{figure}[!htb]
  \centering
  \includegraphics[width=1\linewidth]{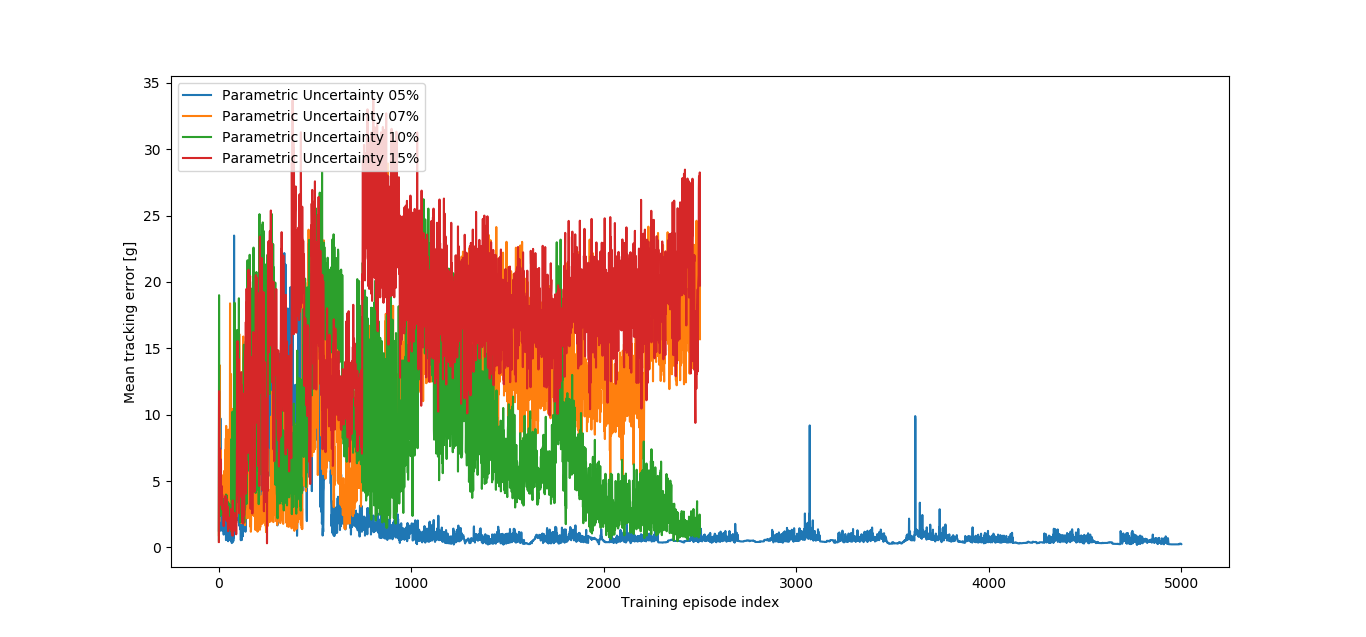}
  \caption[Mean tracking error of parametric uncertainty robustifying trainings]{Mean tracking error of parametric uncertainty robustifying trainings}
  \label{fig:rob_param_trains}
\end{figure}

\begin{figure}[!htb]
    \begin{subfigmatrix}{2}
        \subfigure[Tracking performance]{\includegraphics[width=0.49\linewidth]{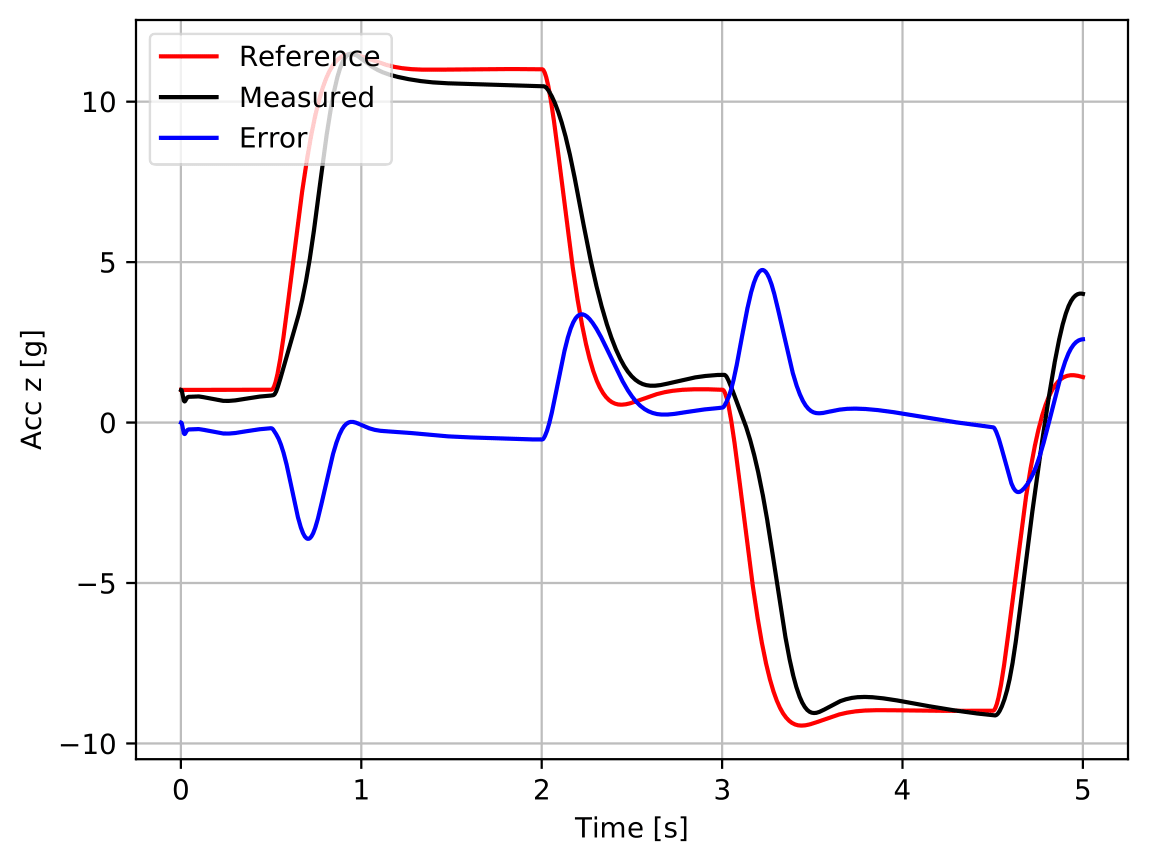}}
        \subfigure[Action, $\eta$]{\includegraphics[width=0.49\linewidth]{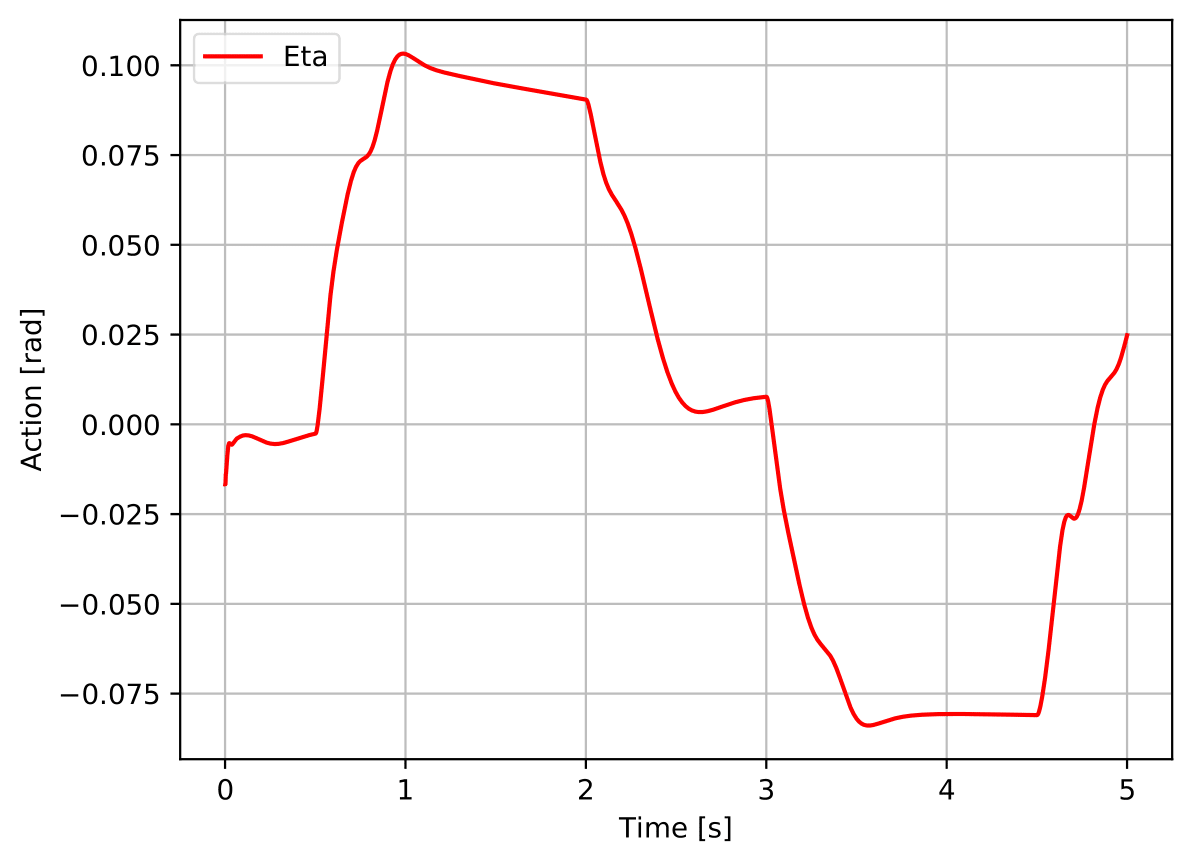}}
    \end{subfigmatrix}
    \caption[Nominal Performance of the Parametric Uncertainty Robustified Agent]{Nominal Performance of the Parametric Uncertainty Robustified Agent}
    \label{fig:nom_test_param_rob_agent}
\end{figure}

As figure \ref{fig:rob_param_trains} shows, from the four different values of $3\sigma$ tried, only the 5\% one converged after 2500 episodes, meaning that it was the only robustifying training being run for a total of 5000 episodes, during which the best agent found was defined as the Parametric Uncertainty Robustified Agent. The nominal tracking performance is damaged, but the action signal is smoother (cf. figure \ref{fig:nom_test_param_rob_agent}).

Furthermore, it enhanced the performance of the Nominal Agent in environments with parametric uncertainty (-40\% to 40\%), having achieved high success rates (cf. table \ref{tab:param_test_comparison}).

\begin{table}[!htb]
  \renewcommand{\arraystretch}{1.2} 
\caption{Robustified Agent success rate in improving the robustness to Parametric Uncertainty of the Nominal Agent}
  \begin{tabular}{lcc}
    \hline
    \textbf{Requirement} &  \textbf{Success \%} \\
    \hline
    $|e_z|_{\max,r}$    &   61.55 \\
    Overshoot   &    83.16  \\
    $\eta_{\max}$  &    98.28  \\
    $\eta_{noise,r}$    &    100.00  \\
    $\eta_{noise,t}$    &    99.31    \\
    \hline
  \end{tabular}
  \label{tab:param_test_comparison}
\end{table}

Since the Parametric Uncertainty Robustified Agent was better than the Nominal Agent in non-nominal environments, it won in the Robustness categories, remaining acceptably the same in terms of performance in nominal environments. Thus, it is possible to say, in general terms, that, concerning parametric uncertainty, the robustifying trainings succeeded.

\section{Achievements}
\label{section:achievements}

The proposed algorithm has been considered successful, since all the objectives established in section \ref{sec:intro} were accomplished, confirming the motivations put forth in section \ref{sec:intro}. Three main achievements must be highlighted:

\begin{enumerate}
    \item the nominal target performance (cf. section \ref{section:optimal_agent}) achieved by the proposed algorithm with the non-linear model of the dynamic system (cf. section \ref{sec:model});
    \item the ability of SER (cf. section \ref{section:SER}) in boosting a previously suboptimally converged performance;
    \item the very sound rates of success in overtaking the performance achieved by the best found nominal agent (cf. sections \ref{section:robustness_estimation_uncertainty} and \ref{section:robustness_parametric_uncertainty}). RL has confirmed to be a promising learning framework for real life applications, where the concept of Robustying Trainings can bridge the gap between training the agent in the nominal environment and deploying it in reality.
\end{enumerate}

\section{Future Work}
\label{section:future}
The first direction of future work is to expand the current task to the control of the whole nonlinear flight dynamics of the GSAM, instead of solely the longitudinal one. Such an expansion would require both (i) straightforward modifications in the code and in the training methodology and (ii) some conceptual challenges, concerning the expansion of the reward function and of the exploration strategy.

Secondly, it would be interesting to investigate how to tackle the main challenges faced during the design of the proposed algorithm, namely (i) to avoid the time-consuming reward engineering process, (ii) the definition of the exploration strategy and (iii) the reproducibility issue.

\bibliographystyle{plain}
\bibliography{00_Manuscript}   

\end{document}